\documentclass[letterpaper]{article} % DO NOT CHANGE THIS
\usepackage{aaai24}  % DO NOT CHANGE THIS
\usepackage{times}  % DO NOT CHANGE THIS
\usepackage{helvet}  % DO NOT CHANGE THIS
\usepackage{courier}  % DO NOT CHANGE THIS
\usepackage[hyphens]{url}  % DO NOT CHANGE THIS
\usepackage{graphicx} % DO NOT CHANGE THIS
\usepackage{natbib}  % DO NOT CHANGE THIS AND DO NOT ADD ANY OPTIONS TO IT
\usepackage{caption} % DO NOT CHANGE THIS AND DO NOT ADD ANY OPTIONS TO IT
\usepackage{algorithm}
\usepackage{algorithmic}

\usepackage{booktabs}
\usepackage{multirow}
\usepackage{amssymb}

\usepackage{newfloat}
\usepackage{listings}

\DeclareCaptionStyle{ruled}{labelfont=normalfont,labelsep=colon,strut=off} % DO NOT CHANGE THIS
\floatstyle{ruled}
\newfloat{listing}{tb}{lst}{}
\floatname{listing}{Listing}
\title{GridFormer: Point-Grid Transformer for Surface Reconstruction}
\author{
    Shengtao Li\textsuperscript{\rm 1,\rm 2}, Ge Gao\textsuperscript{\rm 1,\rm 2}\thanks{Corresponding author: Ge Gao}, Yudong Liu\textsuperscript{\rm 1,\rm 2}, Yu-Shen Liu\textsuperscript{\rm 2}, Ming Gu\textsuperscript{\rm 1,\rm 2}
}
\affiliations{
    \textsuperscript{\rm 1}Beiing National Research Center for Information Science and Technology (BNRist), Tsinghua University, Beijing, China\\
    \textsuperscript{\rm 2}School of Software, Tsinghua University, Beijing, China\\
    list21@mails.tsinghua.edu.cn, gaoge@tsinghua.edu.cn,\\ liuyd23@mails.tsinghua.edu.cn, liuyushen@tsinghua.edu.cn, guming@tsinghua.edu.cn
}

\usepackage{bibentry}
\begin{document}

\maketitle

\begin{abstract}
Implicit neural networks have emerged as a crucial technology in 3D surface reconstruction. To reconstruct continuous surfaces from discrete point clouds, encoding the input points into regular grid features (plane or volume) has been commonly employed in existing approaches. However, these methods typically use the grid as an index for uniformly scattering point features. Compared with the irregular point features, the regular grid features may sacrifice some reconstruction details but improve efficiency. To take full advantage of these two types of features, we introduce a novel and high-efficiency attention mechanism between the grid and point features named Point-Grid Transformer (GridFormer). This mechanism treats the grid as a transfer point connecting the space and point cloud. Our method maximizes the spatial expressiveness of grid features and maintains computational efficiency. Furthermore, optimizing predictions over the entire space could potentially result in blurred boundaries. To address this issue, we further propose a boundary optimization strategy incorporating margin binary cross-entropy loss and boundary sampling. This approach enables us to achieve a more precise representation of the object structure. Our experiments validate that our method is effective and outperforms the state-of-the-art approaches under widely used benchmarks by producing more precise geometry reconstructions. The code is available at https://github.com/list17/GridFormer.
\end{abstract}

\section{Introduction}

Perceiving and modeling the surrounding world are essential tasks in 3D computer vision. Point clouds obtained from various sensors allow us to capture discrete spatial information about 3D surfaces directly. Surface reconstruction plays a vital role in converting this discrete representation into a continuous one. Recently, learning-based approaches have gained significant popularity in reconstructing point clouds. These implicit methods employ a conversion process that transforms the input point cloud into a global feature to represent the spatial continuous field of a 3D shape. However, bridging the gap between the continuous space and the discrete point cloud poses challenges, resulting in varying reconstruction outcomes across different representations.

Typically, these methods encode the point cloud using either a single global feature or regular local grid features. The regular grid features are learned by uniformly distributing the point-wise features across each grid. While regular grid features capture information averagely from every point of the space, they may overlook the shape details in the point cloud. Some other methods will attach the features to the input points \cite{Boulch_2022_CVPR, zhang2022dilg} or move the regular grid features close to the surface \cite{Li2022DCCDIF}. These irregular features can represent the 3D shape more accurately. However, connecting the irregular features with the space can be difficult. Also, the target occupancy function is not differentiable or even not continuous at the zero level. This intrinsic property increases the error bound and makes training more difficult.

\begin{figure}[t]
    \centering
    \includegraphics[width=0.98\linewidth]{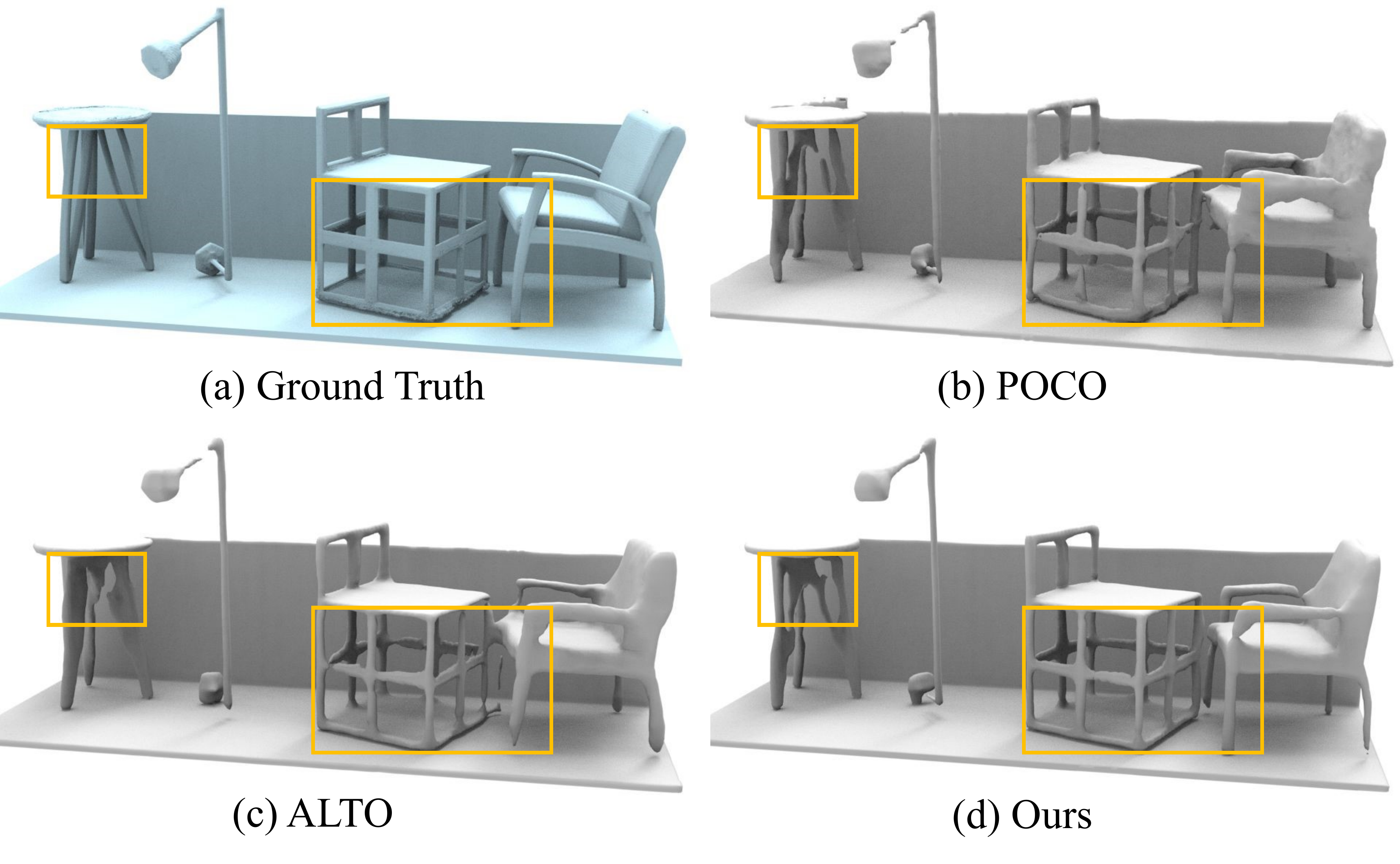}
    \caption{Visualization of the complex scene reconstruction results on the Synthetic Rooms dataset \cite{Peng2020ECCV}. Our method can produce high-fidelity reconstructions compared with the point-based method POCO \cite{Boulch_2022_CVPR} and the grid-based method ALTO \cite{Wang2023CVPR}.}
    \label{fig:teaser}
\end{figure}

\begin{figure}[t]
    \centering
    \includegraphics[width=0.98\linewidth]{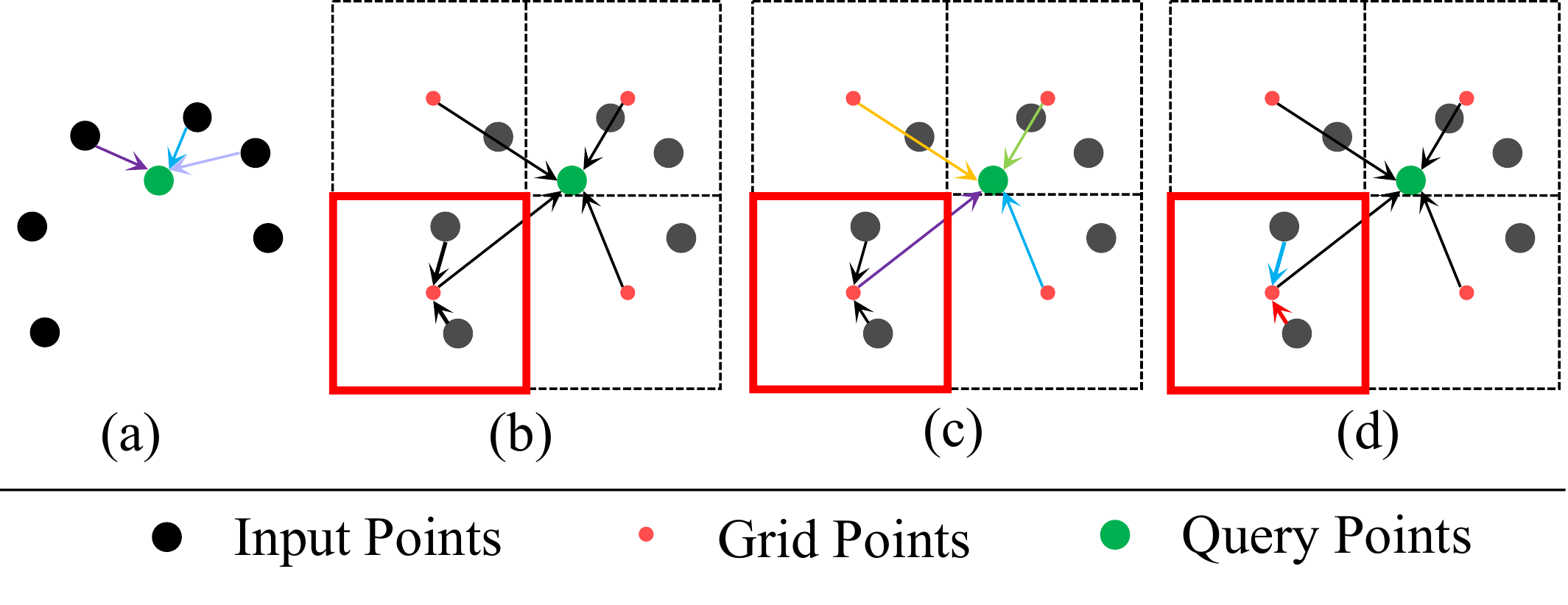}
    \caption{Comparisons between our GridFormer and other methods. The colorized arrows in (a), (c), and (d) represent learnable weights for scattering point or grid features. (a) The point-based approach expresses the query point feature by aggregating the nearby point features with learnable weights. (b) The grid-based approach learns the grid features by uniformly scattering the point features. The decoder aggregates the grid features by the weights calculated by bilinear or trilinear interpolation. (c) The attention-based decoder in ALTO \cite{Wang2023CVPR} makes the weights between the query and grid points learnable. (d) Our point-grid transformer learns the weights between the input and grid features. This enables our method to approximate (a) through grid points while maintaining high efficiency.}
    \label{fig:difference}
\end{figure}

To address these challenges, we propose a novel and highly efficient attention mechanism that bridges the space and point cloud by treating the grid as a transfer point. Figure \ref{fig:difference} shows the difference between our approach and the previous techniques based on point or grid structures. The point-based method can effectively obtain the shape information from the point cloud, but the irregular structure of the points significantly decreases efficiency. Our approach leverages the concept of point-grid attention to model the grid feature. This enables our network to learn the relationship between the input and grid features, which can implicitly bridge the space and the points. And the visual reconstruction differences between these methods can be found in Figure \ref{fig:teaser}.

In particular, apart from employing uniform sampling, we have devised a two-stage training strategy. It incorporates margin binary cross-entropy loss and boundary sampling to narrow down the error bound caused by the discontinuity property, ensuring a more precise reconstruction result.

Our contributions can be listed as follows:
\begin{itemize}
    \item We introduce the Point-Grid Transformer (GridFormer) for surface reconstruction. Our method significantly improves the spatial expressiveness of grid features for learning implicit neural fields.
    \item We design a two-stage training strategy incorporating margin binary cross-entropy loss and boundary sampling. This strategy enhances our model's predictive capability by yielding a more precise occupancy field near the surface.
    \item Both object-level and scene-level reconstruction experiments validate our method, demonstrating its effectiveness and ability to produce accurate reconstruction results.
    
\end{itemize}

\section{Related Work}
\subsection{3D Representations}

\subsubsection{Explicit Representations.} Voxels are amongst the most widely used shape representations \cite{Maturana_Scherer_2015, choy20163d}. However, as the resolution increases, the memory consumed by voxels increases dramatically. Different from voxels, point clouds represent a 3D shape as a set of discrete points \cite{qi2016pointnet, qi2017pointnetplusplus}. These points are irregularly distributed in space and lack continuous topological relationships. Hence post-processing steps \cite{kazhdan2013screened} are needed to extract continuous surface. Meshes \cite{gkioxari2019meshrcnn, pan2019deep} avoid the complexity brought by voxels and can better represent the topological structure. However, generating mesh directly from the neural network is also more complicated. Most meshed-based methods require deforming geometric primitives \cite{groueix2018AtlasNet} or templates of fixed topology \cite{Groueix20183DCODED3C}.
 
\subsubsection{Neural Implicit Representations.}
Implicit representation characterizes the whole space by predicting each point as inside, outside, or on the surface. It relies on a neural network acting as the function to model the binary occupancy field \cite{Chen_Zhang_2019, Mescheder_Oechsle_Niemeyer_Nowozin_Geiger_2019, NEURIPS2019_b5dc4e5d} or distance field \cite{icml2020_2086, Park_2019_CVPR, Atzmon_2020_CVPR, takikawa2021nglod} and then uses the marching cubes \cite{10.1145/37401.37422} algorithm to extract the mesh. The implicit function typically receives a query point with a feature and outputs the corresponding occupancy or distance value. A single global feature has been applied to represent different shapes initially but it cannot capture local details. To resolve this, several works explored capturing local features both in the 2D image field \cite{pifuSHNMKL19, saito2020pifuhd, NEURIPS2019_39059724} and in the 3D point cloud field \cite{chibane20ifnet, Peng2020ECCV, LPI, PredictiveContextPriors}.

To obtain more faithful geometric features, some subsequent works also explored multi-resolution \cite{takikawa2021nglod, Chen2021Multiresolution, huang2023nksr}, irregular or dynamic feature representations \cite{Li2022DCCDIF, Boulch_2022_CVPR, zhang2022dilg}. Indeed, compared to the regular gird features \cite{Peng2020ECCV, tang2021sign, Lionar2021WACV}, the irregular feature representation is more compact and better suited to capture the details. Some other works \cite{NeuralPull, ben2022digs, BaoruiTowards} use the gradient or divergence to constrain the implicit fields for better reconstruction. Due to the intrinsic properties of the occupancy field, we adopt the margin to optimize our estimated occupancy function.

\begin{figure*}[t]
\centering
\includegraphics[width=0.98\textwidth]{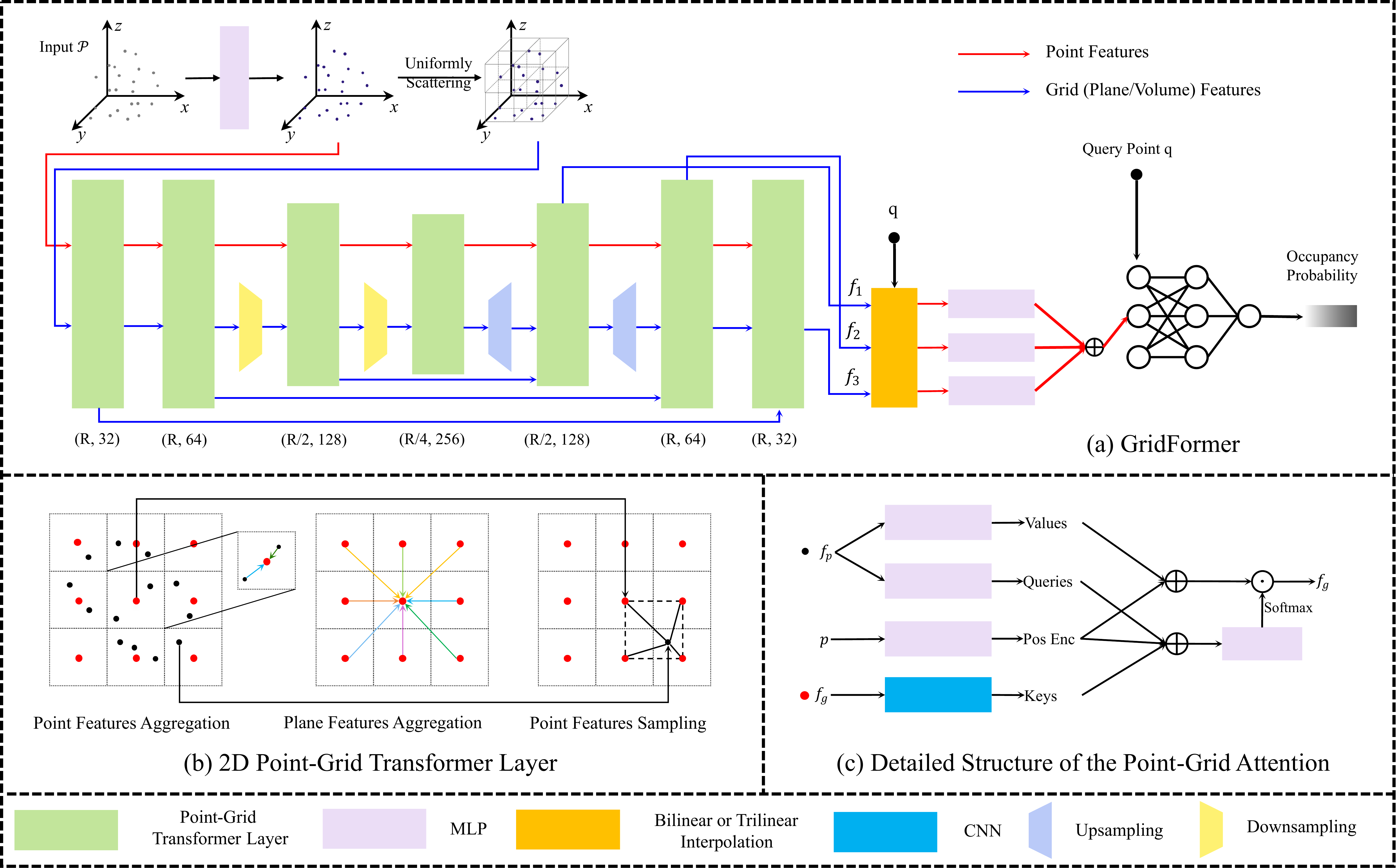} % Reduce the figure size so that it is slightly narrower than the column.
\caption{Overview of our method. (a) The architecture of GridFormer. (b) The 2D plane point-grid transformer layer in which the colorized arrows represent learnable weights. (c) The detailed structure of the point-grid attention mechanism for point features aggregation. `Pos Enc' denotes position encoding.}
\label{ModelOverview}
\end{figure*}

\subsection{Transformers for Point Cloud}
Transformer was first applied in NLP tasks \cite{vaswani2017attention} and has made a great success. It relies on a self-attention mechanism to capture the relationships between different words. This practical approach has also brought about innovations in other fields. Recently several works, like Point Transformer \cite{zhao2021point} and Point Cloud Transformer \cite{Guo_2021}, have explored the application of transformers in point cloud processing. However, due to the discrete distribution of point clouds, many of these approaches rely on the k-nearest neighbor (kNN) search to find the nearest neighboring points. As the size of the point cloud increases, the kNN search becomes more complex and computationally expensive. Fast Point Transformer \cite{park2022fast} designs a lightweight self-attention layer that uses the voxel hashing-based architecture to boost computational efficiency. Our method also utilizes grids for the aggregation of point features. But our innovation lies in utilizing the fixed grid not only for acceleration but also to connect the space and the point cloud.

\section{Method}
Our method aims to establish an efficient attention mechanism for connecting the space and the point cloud. Figure \ref{ModelOverview}(a) shows an overview of our network structure. Based on our point-grid transformer layer, the model constructs a continuous occupancy function $o: \mathbf{R}^3 \rightarrow [0, 1]$. For a point cloud $\mathcal{P} = \{p_i \ |\ p_i \in \mathbf{R}^3\}$, we first learn the per-point features by applying a small point-wise multi-layer perception (MLP). Then the grid (plane or volume) features are initialized by scattering the point features uniformly. The U-Net-like network is constructed based on the point-grid transformer layer. Each layer takes in point and grid features. In the following sections, we will elaborate on the point-grid transformer layer, the multi-resolution decoder, and our subsequent optimization strategy.

\subsection{Point-Grid Transformer Layer}
Given the point cloud $\mathcal{P}$, we define the $f_p, f_g, f_q$ as the features of input, grid, and query points, and the $p, g, q$ represent the corresponding points. $\phi$ and $\psi$ represent the MLP and convolutional neural network (CNN), respectively. The steps of the point-grid transformer layer, as illustrated in Figure \ref{ModelOverview}(b) and (c), will be described in detail below.

\subsubsection{Position Encoding.} We convert the point clouds from the global coordinate system to the local coordinate system of the grid where the point is situated. Then we use an MLP $\phi_{pos}$ with two linear layers and one ReLU nonlinearity function. It takes the localized points as input and outputs the position encoding, as denoted by
\begin{equation}
f_{pos} = \phi_{pos}(p - \lfloor (p \times r) \rfloor / r),
\end{equation}
where $r$ represents the plane or volume resolution.

\subsubsection{Point Features Aggregation.}
This section will describe the point-grid attention mechanism used to aggregate the point features. We leverage the point transformer mechanism in \cite{zhao2021point} to learn the weights between the point and grid features. The detailed structure is shown in Figure \ref{ModelOverview}(c). A small CNN $\psi_k$ is applied to the grid features to learn the keys, and two MLP networks $\phi_q$ and $\phi_v$ are utilized to learn the queries and values of the point features, respectively. We also add position encoding in both the point-grid attention generation branch and the feature transformation branch following \cite{zhao2021point}. The point-grid weights can be represented as follows:
\begin{equation}
    \omega_{ij} = \phi_w(\psi_k(f_{g_i}) - \phi_q(f_{p_j}) + f_{pos}).
\end{equation}
Here the point $p_j$ is situated in the same grid $\mathcal{N}_{g_i}$ as the grid point $g_i$. The softmax function is applied to normalize the weights in the same grid. Then we aggregate the point features by our learned point-grid weights:
\begin{equation}
    f_{g_i}=\sum_{p_j \in \mathcal{N}_{g_i}} \omega_{ij} (\phi_v(f_{p_j}) + f_{pos}).
\end{equation}
\subsubsection{Grid Features Aggregation.}
To increase the receptive field of each grid, we further aggregate the neighbor grid features by a small CNN. Considering our motivation and the lightweight network design, we replace the CNN with the depth-wise convolutional network \cite{chollet2017xceptiondepthwise} in the last three point-grid transformer layers directly connected to the decoder. The depth-wise convolution convolves each input channel with a different kernel which acts as the shared weights between the grid features.

\subsubsection{Point Features Sampling.}
Merely updating the point features from $f_{p_j}$ to $\phi_v(f_{p_j}) + f_{pos}$ will disregard the point features within the neighborhood. To address this issue, we opt to sample the point features from the hybrid grid features using bilinear or trilinear interpolation denoted as $\mathcal{S}$. Simultaneously, considering that the grid features inherently contain position encoding, we omit this component during this stage. The revised point features can be mathematically expressed as
\begin{equation}
    f_{p_i} = \phi_v(f_{p_j}) + \mathcal{S}(f_g).
\end{equation}
Throughout this entire stage, the grid features only serve the purpose of computing the keys. Consequently, we implement a skip connection for both the point and grid features within one attention layer.

\subsection{Multi-resolution Decoder}
For any given query point $q$, we can sample the query feature by bilinear or trilinear interpolation, leveraging our learned grid features as shown in Figure \ref{fig:difference}(d). The weights in the interpolation are fixed for a certain query point since they are computed by the relative distance. An improved method is learning the weights also through the attention mechanism between the query and grid points like the Figure \ref{fig:difference}(c). However, In the feature transfer chain from the input point cloud to the grid points, then to the query points, we already make the weights of the front half part learnable. At the same time, the number of query points will increase rapidly when we need a high-resolution reconstruction result. An attention-based decoder will consume more time to decode the occupancy value for each query point.

Based on the aforementioned factors, we still keep the interpolation method to sample the query features. We opt for grid features $f = (f_1, f_2, f_3)$ from two different resolutions as illustrated in Figure \ref{ModelOverview}(a). We scale the different feature dimensions to 32 through a shallow MLP to keep the same settings as other methods. We use the same network $o_\theta$ for the decoder as \cite{Peng2020ECCV}, which takes in the accumulated feature $f_q \in \mathcal{F}$ and the query point $q \in \mathbf{R}^3$ and outputs the occupancy:
\begin{equation}
    o_\theta : \mathbf{R}^3 \times \mathcal{F} \rightarrow [0, 1].
\end{equation}
\subsection{Boundary Optimization}
\begin{figure}[t]
    \centering
    \includegraphics[width=0.98\linewidth]{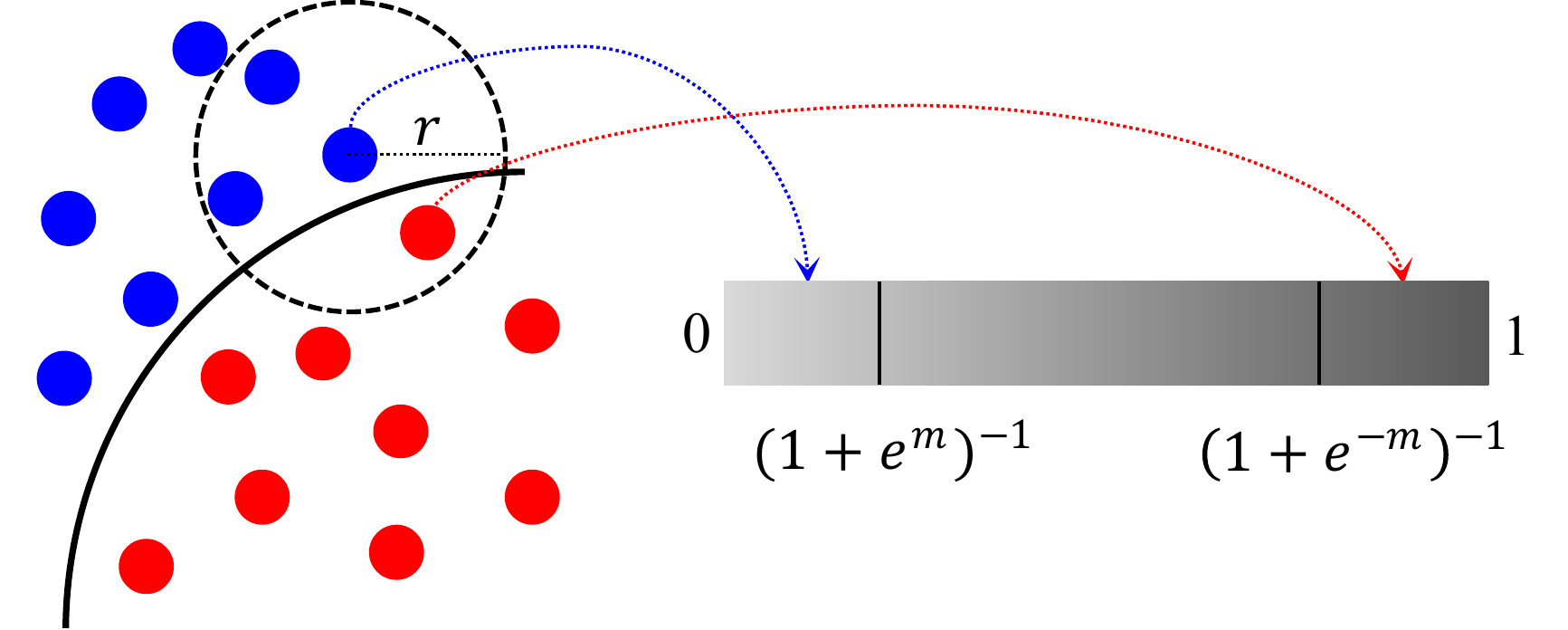}
    \caption{Illustration of boundary optimization.}
    \label{fig:boundary_optimization}
\end{figure}

In this section, we propose boundary optimization for the occupancy function. According to the definition of the occupancy function $o: \mathbf{R}^3 \rightarrow [0, 1]$, it is not continuous and not differential on the surface. Based on these intrinsic features, we adopt the margin binary cross-entropy loss to finetune our model with the boundary sampling.

\begin{table*}[t]
\centering
\resizebox{\textwidth}{!}{
\begin{tabular}{l|cccc|cccc|cccc}
\toprule

\multirow{2}{*}{Methods}      & \multicolumn{4}{c|}{3000 Pts, $\sigma$ = 0.005} & \multicolumn{4}{c|}{1000 Pts, $\sigma$ = 0.005}& \multicolumn{4}{c}{300 Pts, $\sigma$ = 0.005}   \\ 

\cmidrule{2-13}

        & IoU $\uparrow$  & CD $\downarrow$   & NC $\uparrow$    & \multicolumn{1}{c|}{FS $\uparrow$} & IoU $\uparrow$   & CD $\downarrow$  & NC $\uparrow$  & \multicolumn{1}{c|}{FS $\uparrow$} & IoU $\uparrow$ & CD $\downarrow$  & NC $\uparrow$  & FS $\uparrow$  \\ 
        
\midrule
        
ONet \cite{Mescheder_Oechsle_Niemeyer_Nowozin_Geiger_2019}    & 0.761  & 0.87& 0.891  & 0.785  & 0.772  & 0.81& 0.894  & 0.801  & 0.778  & 0.80 & 0.895  & 0.806  \\
ConvONet \cite{Peng2020ECCV}& 0.884  & 0.44& 0.938  & 0.942  & 0.859  & 0.50 & 0.929  & 0.918  & 0.821  & 0.59& 0.907  & 0.883  \\
POCO \cite{Boulch_2022_CVPR}   & 0.926  & 0.30 & 0.950& 0.984  & 0.884  & 0.40 & 0.928  & 0.950   & 0.808  & 0.61& 0.892  & 0.869  \\
ALTO \cite{Wang2023CVPR}   & 0.930  & 0.30 & 0.952  & 0.980   & 0.905  & 0.35& 0.940& 0.964  & 0.863  & 0.47& 0.922  & 0.924  \\ 

\midrule

% Ours (w/o refine)& 0.936  & 0.28& 0.955  & 0.985  &  0.912 & 0.34 & 0.946 & 0.970 & 0.864  & 0.46& 0.924  & 0.926  \\
Ours & \textbf{0.936}  & \textbf{0.28}&  \textbf{0.956} & \textbf{0.985}  & \textbf{0.913}    & \textbf{0.33}     & \textbf{0.946}    &   \textbf{0.970}     & \textbf{0.866}  & \textbf{0.46} &  \textbf{0.925}  & \textbf{0.926}  \\

\bottomrule
\end{tabular}}
\caption{Comparison on the ShapeNet dataset with different point density levels. `Pts' denotes input points and $\sigma$ is the standard deviation of the Gaussian noise.}
\label{shapenet}
\end{table*}

\subsubsection{Boundary Sampling.}
The uniform sampling strategy has been proven to be the most suitable training strategy in \cite{Mescheder_Oechsle_Niemeyer_Nowozin_Geiger_2019} as other alternative sampling strategies tend to introduce bias to the model. But for a precise reconstruction, it is required to have accurate predictions near the surface. Hence, we divide the whole training procedure into two stages. At the first stage, we use uniform sampling for training until the model converges. At the second stage, we switch to boundary sampling. To ensure fairness, we extract boundary regions from the original training data rather than resampling the query points.

Since both the noise levels and densities of the point clouds are different, extracting the region based on the input point cloud is insufficient. We extract the boundary points based on the ground-truth occupancy labels the same as \cite{tang2022cbl}. A point is a boundary point only if at least one of its neighboring points lies on the opposite side of the surface. A fixed radius $r$ is set to search for the opposite points shown in Figure \ref{fig:boundary_optimization}.

\subsubsection{Margin Binary Cross-entropy Loss.}
At the first stage, we minimize the binary cross-entropy loss between the predicted $\hat{o}_q$ and the ground-truth occupancy values $o_q$ with uniformly sampled points $q \in \mathbf{R}^3$:
\begin{equation}
    \mathcal{L}(\hat{o}_q, o_q) = -[o_q \cdot log(\hat{o}_q) + (1 - o_q) \cdot log(1 - \hat{o}_q)]
\end{equation}
where $o_q$ is calculated by applying a sigmoid layer to the output of the network $o(q)$:
\begin{equation}
    o_q = \frac{1}{1+e^{-o(q)}}.
\end{equation}

A margin $m$ is added directly to the output according to the ground-truth label $l \in [0, 1]$. Consequently, the new output is defined as:
\begin{equation}
    o_q = \frac{1}{1+e^{-(o(q) - m \times (l \times 2 - 1))}}.
\end{equation}
As shown in Figure \ref{fig:boundary_optimization}, the margin binary cross-entropy loss can make the predicted occupancy values as close to 0 or 1 as possible.

\subsection{Implementation Details}
We implement our model in Pytorch \cite{NEURIPS2019_bdbca288} and use the Adam optimizer \cite{kingma2014adam}. The learning rate is $10^{-4}$ at the first stage, and $10^{-6}$ at the finetune stage. The depth of our U-Net-like encoder is 4, and we do not downsample or upsample the grid features in the two top levels the same as \cite{Wang2023CVPR}. The radius $r$ to search for the opposite points is set to 0.08 and the margin $m$ is set to 2.0. At reference time, we apply Multiresolution IsoSurface Extraction (MISE) \cite{Mescheder_Oechsle_Niemeyer_Nowozin_Geiger_2019} to obtain the mesh.

\begin{figure}[t]
    \centering
    \includegraphics[width=0.98\linewidth]{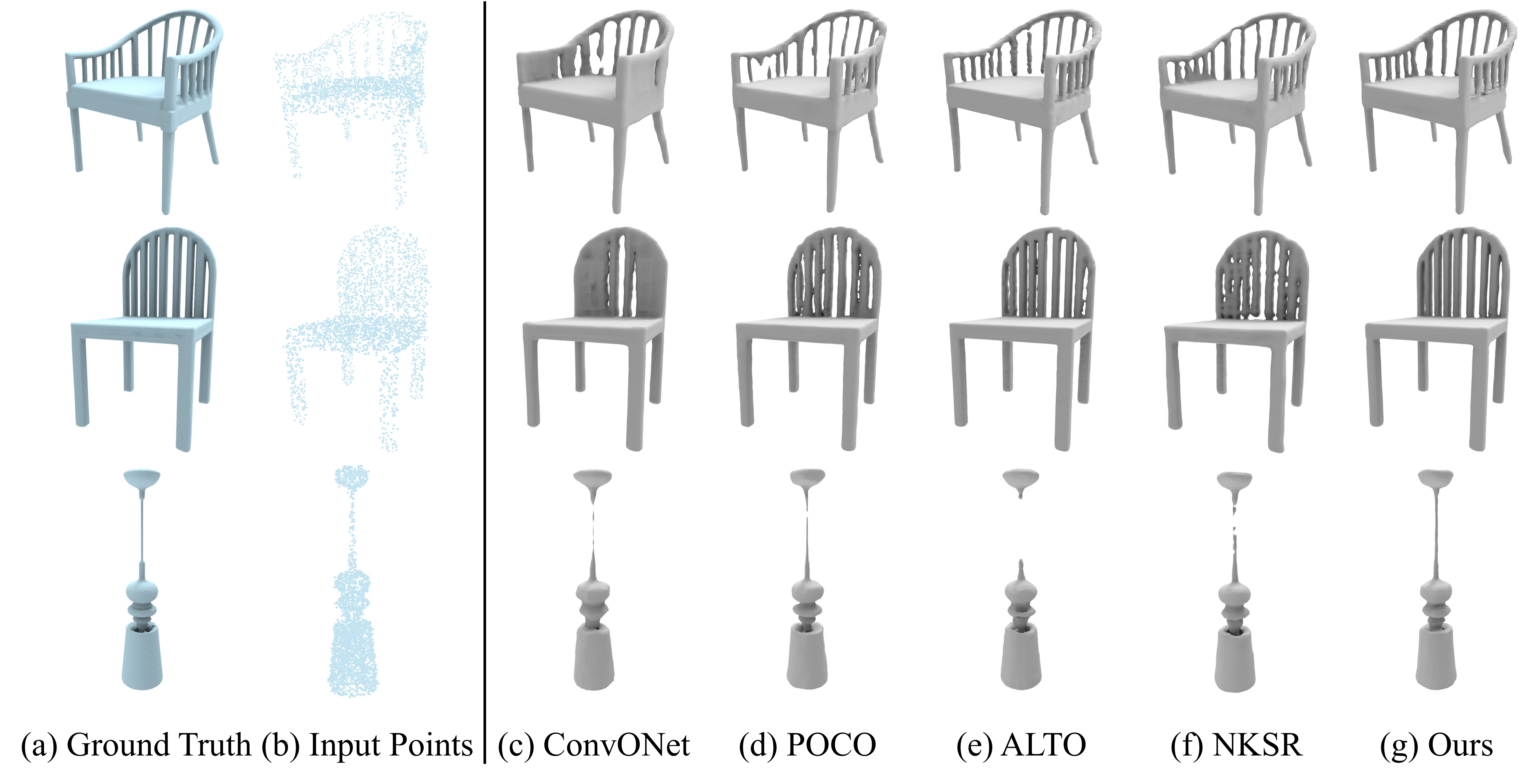}
    \caption{Object-level reconstruction results on the ShapeNet dataset. All the methods are trained and tested on 3000 noisy points.}
    \label{fig:shapnet_results}
\end{figure}

\begin{figure*}[t]
    \centering
    \includegraphics[width=0.98\linewidth]{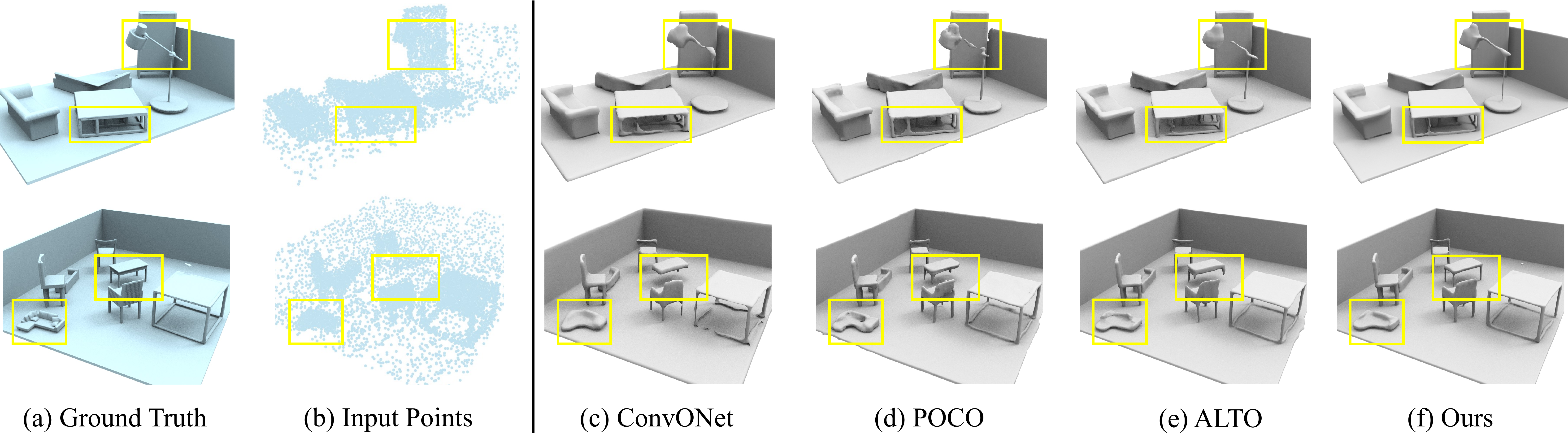}
    \caption{Scene-level comparisons on the Synthetic Rooms dataset. Our method preserves most of the details of the furniture.}
    \label{fig:synthetic_results}
\end{figure*}

\section{Experiments}
\subsection{Datasets, Metrics, and Baselines}

\subsubsection{ShapeNet.}
We use ShapeNet \cite{shapenet2015} for object-level reconstruction evaluation. ShapeNet pre-processed by ONet \cite{Mescheder_Oechsle_Niemeyer_Nowozin_Geiger_2019} contains watertight meshes of shapes in 13 classes, with train/val splits and 8500 objects for testing. To comprehensively evaluate our method, we leverage two different settings for a fair comparison. Following ALTO \cite{Wang2023CVPR}, we sample different densities of points and add Gaussian noise with zero mean and standard deviation of 0.005. To evaluate the effect of noise, we follow NKSR \cite{huang2023nksr} to sample points with different noise levels.

\subsubsection{Scene-Level Datasets.}
For our scene-level reconstruction, we use the Synthetic Rooms dataset \cite{Peng2020ECCV} and ScanNet-v2 \cite{dai2017scannet}. The Synthetic Rooms dataset comprises 5000 synthetic room scenes containing randomly placed walls, floors, and ShapeNet objects. We utilize the same train/validation/test division as previously established.

The ScanNet-v2 dataset includes 1513 scans of real-world environments featuring a diverse selection of room types. The meshes provided in ScanNet-v2 are not watertight, so models are trained using the Synthetic Rooms dataset and then tested on ScanNet-v2. This enables the evaluation of the generalization performance of our method.

\subsubsection{Evaluation Metrics.}
 Following ConvONet \cite{Peng2020ECCV}, we use the volumetric IoU, Chamfer-$L_1$ distance $\times 10^2$ (CD), normal consistency (NC), and F-Score \cite{what3d_cvpr19} with threshold value 1\% (FS) metrics for our evaluation. Other used metrics also include the Chamfer-$L_2$ distance (L2-CD). Please refer to the supplementary of \cite{Peng2020ECCV} for the mathematical details.

\subsubsection{Baselines.}
To evaluate the validity of our attention mechanism, the baselines used for comparison include ONet \cite{Mescheder_Oechsle_Niemeyer_Nowozin_Geiger_2019}, ConvONet \cite{Peng2020ECCV}, POCO \cite{Boulch_2022_CVPR}, and ALTO \cite{Wang2023CVPR}. In addition to these, we also include SPSR \cite{kazhdan2013screened}, SAP \cite{Peng2021SAP}, and NKSR \cite{huang2023nksr}. Please note that NKSR utilized point normals in most of their experiments. To ensure fairness in our comparison, we only evaluate their results from training without point normals.

\begin{table}[h]
\centering
\resizebox{.95\linewidth}{!}{
\begin{tabular}{l|cccccc}
\toprule

\multirow{3}{*}{Methods}     & \multicolumn{2}{c}{1000Pts}  & \multicolumn{2}{c}{3000 Pts}& \multicolumn{2}{c}{3000 Pts}   \\ 
     & \multicolumn{2}{c}{$\sigma$ = 0.0}  & \multicolumn{2}{c}{$\sigma$ = 0.005}& \multicolumn{2}{c}{$\sigma$ = 0.025}   \\ 

\cmidrule{2-7} 

    & IoU $\uparrow$ & CD $\downarrow$ & IoU $\uparrow$ & CD $\downarrow$ & IoU $\uparrow$ & CD $\downarrow$ \\ 
        
\midrule
ConvONet \cite{Peng2020ECCV} & 0.823  & 0.61 & 0.880 & 0.44  & 0.787  & 0.73  \\
SAP \cite{Peng2021SAP}    & 0.908 & 0.34  & 0.911 & 0.33  & 0.829  & 0.53 \\
POCO \cite{Boulch_2022_CVPR} & 0.927  & 0.30 & 0.926 & 0.30  & 0.817  & 0.58 \\
ALTO \cite{Wang2023CVPR} & 0.940  & 0.29 & 0.931 & 0.30  & 0.839  & 0.51 \\ 
NKSR \cite{huang2023nksr}  & 0.934  & \textbf{0.26} & 0.926 & \textbf{0.27}  & 0.829  & \textbf{0.50} \\
\midrule 

% Ours (w/o refine)& 0.945  & 0.28 & 0.936  & 0.28  & 0.844  &0.50    \\
Ours & \textbf{0.946} & 0.28 & \textbf{0.936}  & 0.28  & \textbf{0.844}    & \textbf{0.50} \\

\bottomrule
\end{tabular}}
\caption{Comparison on the ShapeNet dataset with different noise levels.}
\label{shapenet_vary_noise}
\end{table}

\subsection{Object-level Reconstruction}
We first evaluate our method on the task of single-object reconstruction. The quantitative results of different density levels are shown in Table \ref{shapenet}. Our method performs better than the point-based and other grid-based methods. We also find that when the points are too sparse (300 points), the improvement will become smaller because learning the weight for a single point is meaningless. This also verifies the effectiveness of our attention mechanism. We also evaluate the effect of noise following the setting of NKSR \cite{huang2023nksr} in Table \ref{shapenet_vary_noise}. Qualitative comparisons are provided in Figure \ref{fig:shapnet_results}. Compared with other baselines, our method can capture more details, and the overall topology of the shape is more unified and consistent. 

\begin{table}[h]
\centering
\resizebox{0.49\textwidth}{!}{
\begin{tabular}{l|cccc}
\toprule
Methods & IoU $\uparrow$  & CD $\downarrow$   & NC $\uparrow$    & FS $\uparrow$ \\ 
\midrule
        
ONet \cite{Mescheder_Oechsle_Niemeyer_Nowozin_Geiger_2019}    & 0.475  & 2.03& 0.783  & 0.541  \\
SPSR \cite{kazhdan2013screened} & - & 2.23 & 0.866 & 0.810 \\
ConvONet \cite{Peng2020ECCV}& 0.849  & 0.42& 0.915  & 0.964  \\
DP-ConvONet \cite{Lionar2021WACV} & 0.800 & 0.42 & 0.912 & 0.960 \\
POCO \cite{Boulch_2022_CVPR}   & 0.884  & 0.36 & 0.919 & 0.980 \\
ALTO \cite{Wang2023CVPR}   & 0.914  & 0.35 & 0.921  & 0.981   \\ 

\midrule

% Ours (w/o refine)& 0.918  & 0.35& 0.926  & 0.983  \\
Ours & \textbf{0.918} &\textbf{0.34} & \textbf{0.926} & \textbf{0.983} \\

\bottomrule
\end{tabular}}
\caption{Comparison on the Synthetic Rooms dataset.}
\label{tab:synthetic}
\end{table}

\begin{table}[h]
\centering
\resizebox{0.49\textwidth}{!}{
\begin{tabular}{l|cccc}
\toprule
Methods  & CD $\downarrow$   & NC $\uparrow$    & FS $\uparrow$ \\ 
\midrule

ConvONet \cite{Peng2020ECCV} & 0.80& 0.816  & 0.810 \\
DP-ConvONet \cite{Lionar2021WACV} &1.35 &  0.769 & 0.682 \\
POCO \cite{Boulch_2022_CVPR}   & 0.74 & 0.813  &  0.816\\
ALTO \cite{Wang2023CVPR}    & 0.79 & 0.802  & 0.809  \\ 

\midrule

% Ours (w/o refine)& 0.81& 0.844  & 0.786  \\
Ours & \textbf{0.71} & \textbf{0.822} & \textbf{0.846} \\

\bottomrule
\end{tabular}}
\caption{Comparison of the generalization performance on the ScanNet dataset. All methods are trained on the Synthetic Rooms dataset and tested on ScanNet-v2 without floors.}
\label{tab:scannet}
\end{table}

\subsection{Scene-level Reconstruction}
We report numerical comparisons in Table \ref{tab:synthetic} with previous point-based and grid-based methods. The comparison verifies the validity of our proposed mechanism on the scenes. We further present the visual comparison in Figure \ref{fig:synthetic_results}, which shows that our method achieves more detailed reconstruction results, particularly for finer structures.

\begin{figure*}[t]
    \centering
    \includegraphics[width=0.98\linewidth]{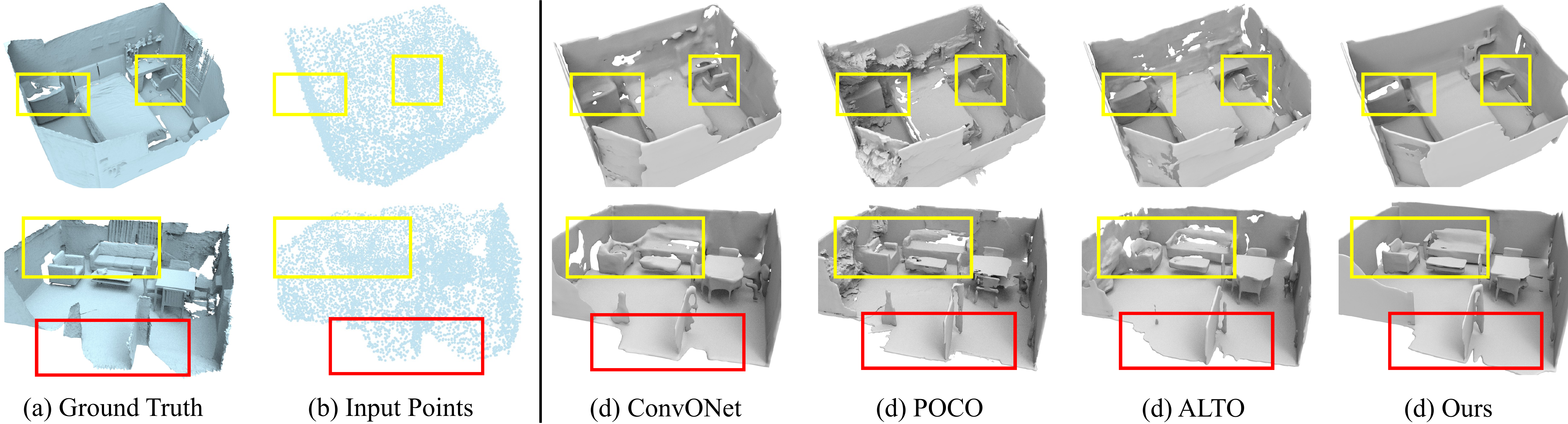}
    \caption{Generalization performance on the ScanNet dataset. These methods are trained on the Synthetic Rooms dataset and tested on the ScanNet-v2 dataset. The red squares show all the methods will complete the floor to some extent.}
    \label{fig:scannet_results}
\end{figure*}

\subsection{Real-world Generalization}
We also explore the generalization performance of our method under the ScanNet-v2 dataset in Table \ref{tab:scannet}. Figure \ref{fig:scannet_results} illustrates that our method can reconstruct a smoother and more complete surface than the other methods. At the same time, we also find that since the Synthetic Rooms dataset only contains regularly generated walls and floors, which are different from the real-scanned rooms as shown in Figure \ref{fig:synthetic_results} and Figure \ref{fig:scannet_results}. This issue causes all methods to try to complete the irregular walls and floors to some extent, like the second group of reconstruction results in Figure \ref{fig:scannet_results}. The completed parts will greatly influence the metrics. For a more accurate comparison, we remove the floors following the method used in ConvONet \cite{Peng2020ECCV} but keep the walls since they are randomly seated throughout the scene.

\subsection{Ablation Study}
\subsubsection{Grid Representation.} We report the effect of our point-grid attention for different representations (triplane and volume) in table \ref{Tab:ablation_structure} on the Synthetic Rooms dataset. All methods use the same decoder without attention. 

\begin{table}[h]
\centering
\resizebox{0.49\textwidth}{!}{
\begin{tabular}{l|cccc}
\toprule
Method  & IoU $\uparrow$  & CD $\downarrow$   & NC $\uparrow$    & FS $\uparrow$ \\ 
\midrule

ConvONet ($3 * 128^2$) \cite{Peng2020ECCV} & 0.805  & 0.44 & 0.903  & 0.948  \\
ConvONet ($64^3$) \cite{Peng2020ECCV}  & 0.849 & 0.42 & 0.915 & 0.964 \\
ALTO ($3 * 128^2$) \cite{Wang2023CVPR} & 0.834 & 0.43 & 0.906 & 0.960 \\
ALTO ($64^3$) \cite{Wang2023CVPR} & 0.903 & 0.36 & 0.920 & 0.981 \\

\midrule
Ours ($3 * 128^2$) & 0.835 & 0.40 & 0.900 & 0.949 \\
Ours ($64^3$) & \textbf{0.918} & \textbf{0.34}  & \textbf{0.926} & \textbf{0.983} \\

\bottomrule

\end{tabular}}
\caption{Ablation on the network framework.}
\label{Tab:ablation_structure}
\end{table}

\subsubsection{Grid Downsampling.} We experiment without downsampling in our U-Net-like encoder and the results are shown in Table \ref{Tab:ablation_downsample}. It takes less time to encode the input points but the reconstruction results are significantly inferior. Thus we consider it worthy to consume a little bit more time for a better result.

\begin{table}[h]
\centering
\resizebox{0.49\textwidth}{!}{
\begin{tabular}{l|cccc}
\toprule
Method  &  IOU $\uparrow$ & CD $\downarrow$ & NC $\uparrow$ & Encode Time (s) $\downarrow$   \\ 
\midrule
Ours (w/o downsampling) & 0.897 & 0.38 & 0.949 & \textbf{0.39}\\
Ours (w/ downsampling)&  \textbf{0.942} & \textbf{0.30} & \textbf{0.962}  & 0.41 \\
\bottomrule

\end{tabular}}
\caption{Ablation on the gird downsampling.}
\label{Tab:ablation_downsample}
\end{table}

\subsubsection{Boundary Optimization.} 
\begin{figure}[h]
    \centering
    \includegraphics[width=0.98\linewidth]{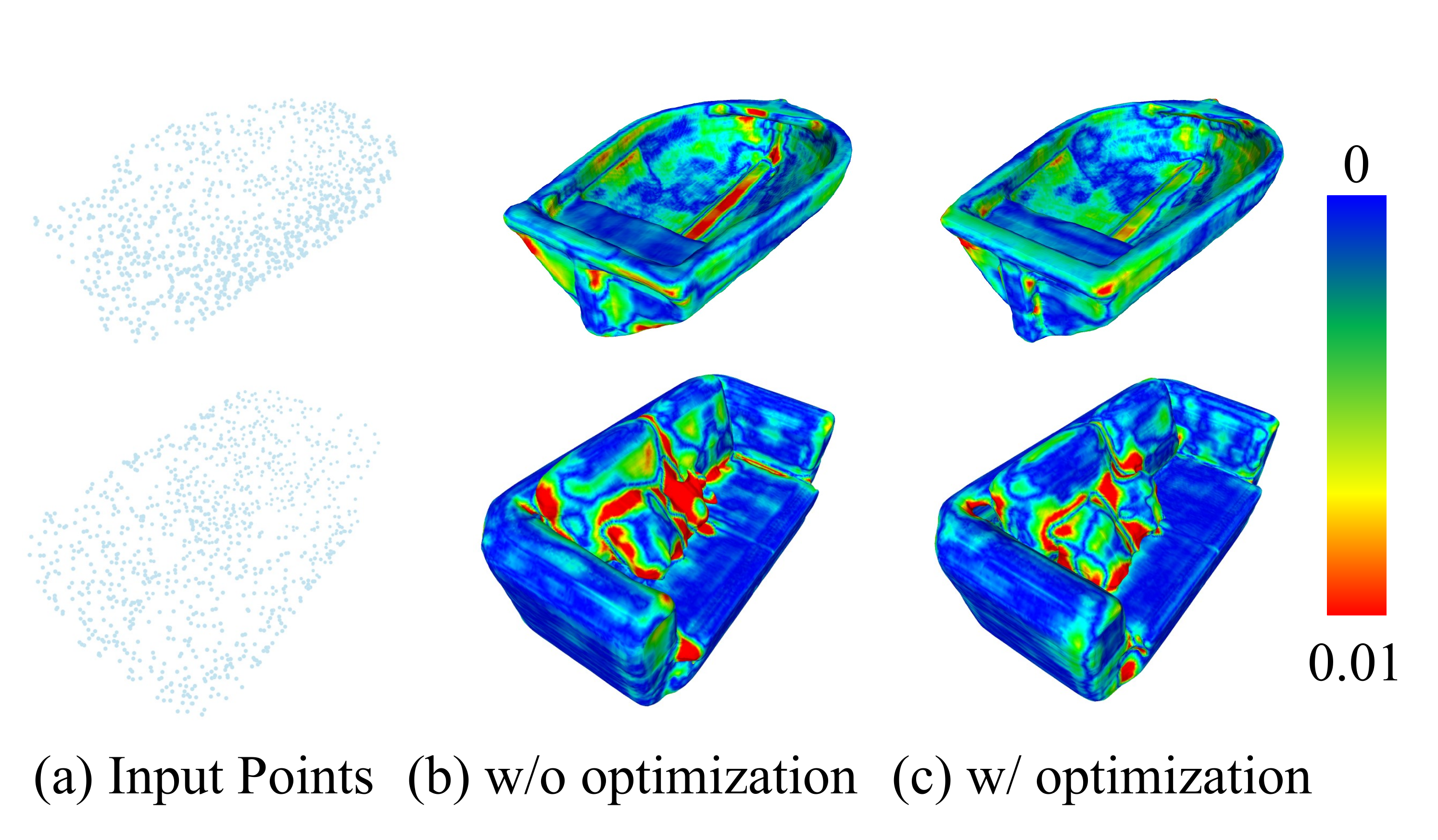}
    \caption{The visual effect of boundary optimization.}
    \label{fig:ablation2}
\end{figure}

We further explore the effect of our boundary optimization. Table \ref{Tab:ablation_optimization} shows that the optimization works under different noise and density levels. We also visualize the distance from reconstruction results to the ground-truth meshes in Figure \ref{fig:ablation2}. The proposed boundary optimization can effectively help reduce the error bound.

\begin{table}[h]
\centering
\resizebox{0.49\textwidth}{!}{
\begin{tabular}{l|cccc}
\toprule
  & 3000 Pts  & 3000 Pts  &  1000 Pts & 1000 Pts \\   
  & $\sigma$ = 0.005  & $\sigma$ = 0.025  &  $\sigma$ = 0   & $\sigma$ = 0.005 \\ 
\midrule
w/o optimzation& 0.211  & 0.881 & 0.253  & 0.371  \\
w/ optimization   & \textbf{0.201}  & \textbf{0.718}  & \textbf{0.247} & \textbf{0.328} \\
\bottomrule

\end{tabular}}
\caption{Ablation study on boundary optimization in terms of L2-CD ($\times 10^4$) .}
\label{Tab:ablation_optimization}
\end{table}

\begin{table}[h]
\centering
\resizebox{0.49\textwidth}{!}{
\begin{tabular}{l|cccc}
\toprule
Method  & GPU Memory (MiB)  & Inference Time (s) \\ 
\midrule

ConvONet \cite{Peng2020ECCV} & 1957  & 0.43   \\
POCO \cite{Boulch_2022_CVPR}  & 6540  & 9.30\\
ALTO \cite{Wang2023CVPR} & 3257  & 7.96 \\
Ours & 1915 &  5.61\\
\bottomrule

\end{tabular}}
\caption{GPU memory and runtime comparisons on ShapeNet chairs.}
\label{Tab:runtime}
\end{table}

\section{Conclusion}
We proposed the Point-Grid Transformer (GridFormer) using a novel point-grid attention mechanism between the point and grid features. It is valid both for object-level and scene-level reconstruction and reconstructs a smoother surface on the unseen dataset. Compared with the attention-based decoder used in the point-based and other grid-based methods, our attention-based encoder costs less time and GPU memory (Table \ref{Tab:runtime}) and achieves comparable or even better results. Our introduced boundary optimization strategy can reduce the error between the estimated and ground-truth occupancy functions and help extract a more accurate surface.

Finally, the experiments show that both the density of the input points and the grid size impact our method's effect. In future work, exploring how to dynamically divide the grid to achieve the attention mechanism between different resolutions may apply this mechanism to more scenarios.

\section*{Acknowledgments}
The corresponding author is Ge Gao. This work was supported by the National Key Research and Development Program of China (2021YFB1600303).

\bibliography{aaai24}

\end{document}